\documentclass[journal]{IEEEtran}
\usepackage{amsmath,amsfonts}
\usepackage{algorithm}
\usepackage{algpseudocode}
\usepackage{amsmath}
\usepackage{array}
\usepackage[caption=false,font=normalsize,labelfont=sf,textfont=sf]{subfig}
\usepackage{textcomp}
\usepackage{stfloats}
\usepackage{url}
\usepackage{verbatim}
\usepackage{graphicx}
\usepackage{cite}
\usepackage[hidelinks,colorlinks=false]{hyperref}
\usepackage{color}
\usepackage{booktabs}

\usepackage{multirow}
\usepackage{diagbox}
\usepackage{scalerel}
\usepackage{tikz}
\usepackage{siunitx}
\usetikzlibrary{svg.path}

\definecolor{orcidlogocol}{HTML}{A6CE39}
\tikzset{
  orcidlogo/.pic={
    \fill[orcidlogocol] svg{M256,128c0,70.7-57.3,128-128,128C57.3,256,0,198.7,0,128C0,57.3,57.3,0,128,0C198.7,0,256,57.3,256,128z};
    \fill[white] svg{M86.3,186.2H70.9V79.1h15.4v48.4V186.2z}
                 svg{M108.9,79.1h41.6c39.6,0,57,28.3,57,53.6c0,27.5-21.5,53.6-56.8,53.6h-41.8V79.1z M124.3,172.4h24.5c34.9,0,42.9-26.5,42.9-39.7c0-21.5-13.7-39.7-43.7-39.7h-23.7V172.4z}
                 svg{M88.7,56.8c0,5.5-4.5,10.1-10.1,10.1c-5.6,0-10.1-4.6-10.1-10.1c0-5.6,4.5-10.1,10.1-10.1C84.2,46.7,88.7,51.3,88.7,56.8z};
  }
}

\newcommand\orcidicon[1]{\href{https://orcid.org/#1}{\mbox{\scalerel*{
\begin{tikzpicture}[yscale=-1,transform shape]
\pic{orcidlogo};
\end{tikzpicture}
}{|}}}}

\begin{document}

\title{Sensor-Informed Per-Point Covariance for Structured-Light 3D Imaging}

\author{Sehoon Tak$^{\textsuperscript{\orcidicon{0009-0006-4947-2051}}}$,
and Jae-Sang Hyun$^{\textsuperscript{\orcidicon{0000-0003-1711-8243}}}$

\thanks{Corresponding author: Jae-Sang Hyun}
\thanks{Sehoon Tak and Jae-Sang Hyun are with the Department of Mechanical Engineering, Yonsei University, Seoul 03722, South Korea. Jae-Sang Hyun is also with the Yonsei Institute for Embodied Intelligence, Yonsei University, Seoul 03722, South Korea. (e-mail: hyun.jaesang@yonsei.ac.kr).}}



\maketitle
\begin{abstract}
Per-point uncertainty models are important in structured-light 3D
reconstruction for probabilistic registration, fusion, and quality assessment.
In practice, however, point-cloud covariances are often modeled as isotropic
constants or inferred from local surface geometry and therefore do not
explicitly reflect the measurement process. This is a limitation in fringe
projection profilometry (FPP), where phase noise propagates through calibrated
reconstruction and produces strongly anisotropic 3D uncertainty. This paper
presents a sensor-informed first-order method for constructing a per-point
$3\times3$ covariance field from experimentally measured phase precision and
calibrated phase-to-depth and phase-to-3D mappings. The formulation separates 
a rank-1 phase-induced covariance from an effective full-rank completion obtained 
by incorporating fitted lateral image-space perturbation scales. Repeated-plane experiments under fixed imaging conditions show close alignment of the dominant covariance direction with the viewing ray, and consistency between the dominant 
phase-induced uncertainty scale and scalar depth uncertainty. In G-ICP registration, the proposed covariance substantially improves over a constant isotropic model while providing a sensor-derived uncertainty representation complementary to conventional geometry-based covariances.
\end{abstract}

\begin{IEEEkeywords}
structured light, fringe projection profilometry, uncertainty propagation,
per-point covariance, probabilistic registration, G-ICP
\end{IEEEkeywords}

\section{Introduction}
\IEEEPARstart{S}{tructured-light} 3D reconstruction is widely used in
industrial inspection, biomedical imaging, and robotic perception because it
provides dense and accurate surface measurements over a wide range of scales
\cite{Geng2011StructuredLightTutorial,Salvi2010StructuredLightPatterns,Zuo2018PSPReview}.
As these systems are used increasingly in registration, fusion, and
uncertainty-aware decision making, reporting only nominal point coordinates is
no longer sufficient. Practical reconstruction pipelines also require a
principled per-point uncertainty representation
\cite{Molimard2013UncertaintyFringeProjection,ODowd2021HeightUncertaintyPDF,Segal2009GICP}.
In Gaussian point-based formulations, this uncertainty is naturally represented
by a $3\times3$ covariance matrix for each reconstructed point, which enables
anisotropic confidence ellipsoids and Mahalanobis-weighted estimation
\cite{Segal2009GICP,Huang2021PointCloudRegistrationSurvey}.

In many registration pipelines, however, per-point covariances are not derived
from the sensing process. Instead, they are often modeled as isotropic
constants or estimated from local surface neighborhoods, for example by
assigning small variance along an estimated normal direction and larger
variances in tangential directions
\cite{Segal2009GICP,Huang2021PointCloudRegistrationSurvey}. Such constructions
encode local geometric structure effectively, but they do not explicitly
describe measurement uncertainty.

This distinction matters in fringe projection profilometry (FPP). A projector
emits phase-shifted fringe patterns, a camera records the deformed fringes, and
phase retrieval yields an absolute phase map that is monotonically related to
depth along each camera ray
\cite{Zuo2018PSPReview,Zhang2018AbsolutePhaseReview}. Because phase errors
perturb range much more strongly than lateral position, structured-light
measurements exhibit pronounced axial--lateral anisotropy
\cite{Liu2003CalibrationBasedPhaseShifting,Molimard2013UncertaintyFringeProjection}.
This anisotropy arises primarily from sensing geometry and the phase
measurement process rather than from local surface smoothness.

Prior analyses of structured-light uncertainty often focus on scalar depth
precision by relating phase noise to range error through system geometry and
phase sensitivity
\cite{Liu2003CalibrationBasedPhaseShifting,Molimard2013UncertaintyFringeProjection}.
While valuable, such analyses do not directly provide the full per-point
covariance required for probabilistic 3D registration. A remaining gap is a
practical connection between experimentally measured phase precision,
calibrated reconstruction, and the $3\times3$ covariance used in
uncertainty-aware estimation.

This connection can be established through the calibrated relationship between
absolute phase and depth. Many modern structured-light pipelines avoid an
explicit geometric projector model and instead estimate a per-pixel or locally
parameterized phase-to-depth mapping that absorbs projector non-idealities,
lens distortion, and system geometry into a calibrated model
\cite{Huang2020PhaseDepthCalibration,Feng2021CalibrationReview,Son2024QuasiCalibration}.
When phase precision is characterized experimentally, this mapping provides a
direct route for propagating phase uncertainty into 3D covariance without
relying on local surface neighborhoods.

More broadly, uncertainty evaluation in metrology is often framed through the
GUM law of propagation of uncertainty and, when linearization is inadequate,
through Monte Carlo propagation of distributions 
\cite{jcgm100_2008,jcgm101_2008}. The present work does not attempt a complete
GUM-style uncertainty budget for all system parameters. Instead, it develops a
sensor-informed first-order covariance model from measured phase precision and
calibrated reconstruction. In practical optical-metrology settings, pointwise
measurement capability is also shaped by spatial-resolution limits, viewing
geometry, and surface-dependent effects, including pointwise
accuracy--precision tradeoffs, surface-normal sensitivity, and the validity
limits of linearized transfer-function descriptions
\cite{Hinz2021FPPProduction,Dickins2020MultiViewFPP,Kuehmstedt2009SurfaceNormal,Zhang2018ITFFringeProjection}.

In this work, we propose a sensor-informed first-order covariance model for
structured-light reconstruction by propagating experimentally measured phase
precision through calibrated phase-to-depth and phase-to-3D mappings. The
formulation distinguishes a physically primary phase-induced covariance from a
practical full-rank completion used for downstream registration. The proposed
covariance is independent of local surface neighborhoods and remains directly
interpretable in terms of phase precision, reconstruction sensitivity, and
imaging geometry. To measure phase precision reproducibly, the experiments are
conducted under a controlled operating condition with approximately constant-SNR. We then evaluate the resulting covariance field in generalized ICP
(G-ICP) registration against a constant isotropic model and a conventional
geometry-derived covariance model.

The main contributions of this paper are as follows:
\begin{enumerate}
    \item We formulate a sensor-informed per-point covariance model for
    structured-light reconstruction by propagating experimentally measured
    phase variance through calibrated phase-to-depth and phase-to-3D mappings,
    without relying on local surface neighborhood geometry. The formulation
    distinguishes a physically primary phase-induced covariance from a
    practical full-rank completion used in probabilistic registration.
    \item We characterize the resulting covariance field and show that it
    captures the characteristic axial--lateral anisotropy of structured-light
    measurements in terms of measured phase precision, reconstruction
    sensitivity, and imaging geometry.
    \item We evaluate the proposed covariance field in downstream
    uncertainty-aware rigid registration using G-ICP, and compare it against a
    constant isotropic model and a conventional geometry-derived covariance
    model under controlled structured-light imaging conditions.
\end{enumerate}

\begin{figure*}[t]
\centering
\includegraphics[width=\linewidth]{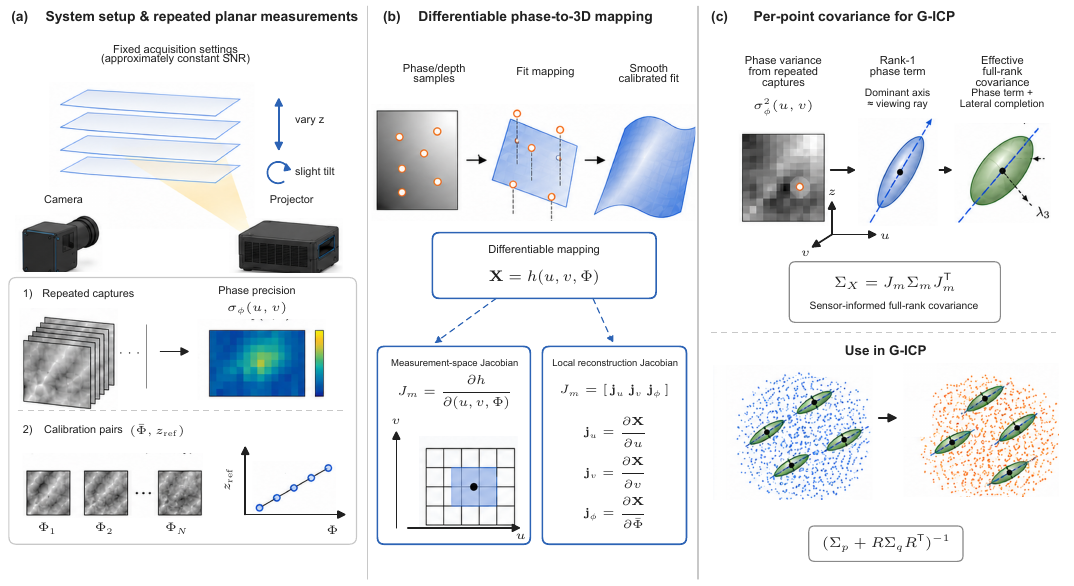}
\caption{Overview of the proposed sensor-informed covariance pipeline for
structured-light reconstruction. (a) Phase precision and phase--depth
calibration data are obtained from repeated planar measurements under an
approximately constant-SNR operating condition. (b) A smooth calibrated
mapping provides local reconstruction sensitivities. (c) Phase uncertainty is
propagated from a rank-1 phase-induced covariance to an effective full-rank per-point covariance, which is subsequently characterized and evaluated in downstream G-ICP registration.}
\label{fig:pipeline}
\end{figure*}

Figure~\ref{fig:pipeline} summarizes the overall workflow of the proposed
method, from phase-precision measurement and phase-to-depth calibration to
covariance construction and downstream evaluation in G-ICP registration.

The remainder of this paper is organized as follows.
Section~\ref{sec:principles} reviews the measurement model, calibrated
reconstruction, and registration background needed for the method.
Section~\ref{sec:covariance} presents the covariance formulation and its
numerical implementation. Section~\ref{sec:metrology} describes the
experimental protocol and implementation details. Section~\ref{sec:results}
reports phase-variance measurements, mapping residuals, covariance
characterization, and registration results. Section~\ref{sec:discussion}
discusses limitations and extensions, including spatially varying SNR and
hybrid sensor--geometry covariance models.

\section{Measurement, Reconstruction, and Registration Background}
\label{sec:principles}

This section reviews the measurement and reconstruction ingredients needed for the proposed covariance model: phase retrieval and phase precision in fringe projection profilometry (FPP), calibrated phase-to-depth and phase-to-3D mapping, coordinate conventions, and the generalized ICP (G-ICP) formulation used for downstream evaluation. The covariance derivation itself is deferred to Section~\ref{sec:covariance}.

\subsection{Absolute phase retrieval and phase precision}
\label{subsec:absolute_phase}

We consider a standard fringe projection profilometry system in which a
projector emits phase-shifted sinusoidal fringe patterns and a camera records
the deformed intensity images. For an $N$-step phase-shifting scheme, the
captured intensity at camera pixel $(u,v)$ is modeled as
\begin{equation}
\label{eq:nstep_psp}
I_n(u,v)=I_0(u,v)+I_m(u,v)\cos\!\big(\phi(u,v)-2\pi n/N\big),
\end{equation}
where $n\in\{0,\ldots,N-1\}$ is the phase-shift index, $I_0(u,v)$ denotes the
average intensity, $I_m(u,v)$ denotes the fringe modulation, and $\phi(u,v)$ is
the wrapped phase \cite{Zuo2018PSPReview}.

Using the standard $N$-step phase-shifting formulation, the wrapped phase is
estimated as
\begin{equation}
\label{eq:wrapped_phase}
\phi(u,v)=
\tan^{-1}\!
\frac{\sum_{n=0}^{N-1} I_n(u,v)\sin(2\pi n/N)}
     {\sum_{n=0}^{N-1} I_n(u,v)\cos(2\pi n/N)}.
\end{equation}
Because the recovered phase is modulo $2\pi$, Gray-code-assisted temporal phase unwrapping is used to obtain the absolute phase $\Phi(u, v)$ \cite{Zuo2016TPUReview,ZHANG2010149}. The subsequent uncertainty analysis is restricted to valid regions without detected fringe-order errors, so that the repeated variability of $\Phi(u, v)$ can be treated as the phase-precision input to the covariance model.

Under fixed acquisition conditions, phase precision is defined as the standard
deviation of repeated absolute-phase measurements at a fixed pixel. Noise-induced
phase error in phase-shifting profilometry depends on image noise, fringe modulation,
and the phase-shifting design \cite{Surrel1997AdditiveNoise, Wang2022NoisePhaseError}.
Practical FPP systems can exhibit systematic phase error arising from nonlinear or nonsinusoidal fringe response, for which dedicated compensation methods have been developed \cite{Wang2021TripleNStep}. The present work characterizes repeatable phase
variance rather than separately modeling individual deterministic error components. Spatial
correlations between neighboring phase errors are neglected in the present first-order
formulation.

\subsection{Calibrated phase-to-depth and phase-to-3D mapping}
\label{subsec:phase_depth_mapping}

A calibrated FPP system requires a deterministic mapping from camera pixel coordinates and absolute phase to 3D geometry. Following common structured-light calibration practice, we separate this mapping into a phase-to-depth relationship and a 3D reconstruction map \cite{Feng2021CalibrationReview,Huang2020PhaseDepthCalibration}. 
We write
\begin{equation}
\label{eq:phase_depth_general}
z = f(\Phi;u,v),
\end{equation}
where $z$ denotes the depth coordinate in the calibrated camera frame and
$f(\cdot)$ is estimated from phase-to-depth calibration data.

Existing phase-to-depth calibration methods include pixel-dependent and
pixel-independent formulations, as well as more flexible pixel-wise or locally
parameterized mappings 
\cite{Pei2022PhaseCoordinates,Zhang2024VirtualPlanes,Son2024QuasiCalibration}. These calibrated models are attractive because they absorb projector non-idealities, distortion, and departures from ideal geometry into the fitted mapping.

In the present work, the calibrated relationship is represented by a smooth
fitted mapping so that local sensitivities can be evaluated reliably. In the
simplest case, the phase-to-depth relationship can be modeled as locally
affine,
\begin{equation}
\label{eq:affine_phase_depth}
z(u,v)=A\,\Phi(u,v)+B(u,v),
\end{equation}
where $A$ is approximately constant over the field and $B(u,v)$ varies slowly
with image position. To account for weak departures from this behavior caused
by distortion or non-ideal geometry, a weakly projective form can also be
used,
\begin{equation}
\label{eq:projective_phase_depth}
z(u,v)=
\frac{A\,\Phi(u,v)+B(u,v)}
     {1+\rho\big(C\,\Phi(u,v)+D(u,v)\big)},
\end{equation}
where $A$ and $C$ are scalar coefficients, $B(u,v)$ and $D(u,v)$ are slowly
varying functions of pixel position, and $\rho$ is chosen so that the
denominator remains close to unity over the calibrated range.

The depth value is then converted to a 3D point through a calibrated
phase-to-3D mapping,
\begin{equation}
\label{eq:phase_to_3d_mapping}
X=
\begin{bmatrix}
x\\y\\z
\end{bmatrix}
=h(u,v,\Phi).
\end{equation}
Equation~\eqref{eq:phase_to_3d_mapping} is written in this general form to
accommodate either an analytic camera model or a smooth calibrated
phase-to-3D mapping. This distinction matters because the covariance model in
Section~\ref{sec:covariance} relies on local differentiation of the calibrated
mapping.

\subsection{Coordinate system and measurement variables}
\label{subsec:coords_measurement}

All reconstructed 3D points are expressed in the reference camera coordinate
frame $C$. The image coordinates $(u,v)$ are defined on the undistorted
reference-camera pixel grid, and the depth variable $z$ denotes the
$z$-coordinate in frame $C$. Under this convention, the reconstruction mapping
$h(u,v,\Phi)$ returns 3D coordinates in metric units.

For later uncertainty propagation, we use the local measurement state $m=[u,\ v,\ \Phi]^\top$, from which the reconstructed point is obtained through $h$. Uncertainty 
in these variables is propagated through the local Jacobian of the calibration mapping
in Section~\ref{sec:covariance}.

\subsection{Probabilistic registration with G-ICP}
\label{subsec:gicp_principles}

To assess the downstream utility of the proposed covariance field, we use the
generalized iterative closest point (G-ICP) framework for rigid point-cloud
registration \cite{Segal2009GICP}. For a correspondence between target point
$p_i$ and transformed source point $Rq_{\pi(i)}+t$, the objective can be written as a
Mahalanobis-weighted residual of the form
\begin{equation}
\label{eq:gicp_cost_compact}
E(R,t)=\sum_i r_i^\top C_i^{-1} r_i,
\end{equation}
\begin{align}
r_i &= p_i-(Rq_{\pi(i)}+t),\\
C_i &= \Sigma_{p_i}+R\Sigma_{q_{\pi(i)}}R^\top.
\end{align}

Conventional G-ICP commonly derives per-point covariances from local surface 
neighborhoods so that they encode local geometric structure\cite{Segal2009GICP}. 
This geometry-derived construction is used here as a registration baseline rather
than as an estimate of the same quantity as the proposed sensor-informed covariance.
Because the Mahalanobis objective requires a numerically invertible covariance, the
downstream application also motivates the effective full-rank covariance developed in Section~\ref{sec:covariance}.

\section{Sensor-informed Per-Point Covariance Formulation}
\label{sec:covariance}

Section~\ref{sec:principles} introduced the ingredients used by the proposed
uncertainty model: measured phase precision, a calibrated phase-to-depth
relationship, and a differentiable phase-to-3D mapping. This section uses
those ingredients to construct a per-point $3\times3$ covariance field for
structured-light reconstruction in the reference camera coordinate frame $C$.
The formulation proceeds in three steps: phase precision is related to depth
uncertainty through the calibrated phase-to-depth mapping in
Eq.~\eqref{eq:phase_depth_general}, the resulting phase-induced uncertainty is
propagated to 3D under a fixed-pixel model, and an effective full-rank
covariance is formed for practical use in probabilistic registration.

\subsection{Covariance construction framework}
\label{subsec:cov_framework}

Using the calibrated phase-to-depth mapping in
Eq.~\eqref{eq:phase_depth_general} and the calibrated phase-to-3D mapping in
Eq.~\eqref{eq:phase_to_3d_mapping}, the reconstruction model is written in
terms of a scalar phase-to-depth transfer and a local phase-to-3D mapping. Here,
$f(\Phi;u,v)$ denotes the calibrated phase-to-depth transfer at camera pixel
$(u,v)$, whereas $h(u,v,\Phi)$ denotes the corresponding local phase-to-3D
reconstruction map used for covariance propagation. Depending on the
implementation, $h$ may be available either directly or through the
composition of the calibrated phase-to-depth mapping with a depth-to-3D
reconstruction model.

We distinguish two covariance quantities. The first is the
\emph{phase-induced covariance}, obtained by propagating phase uncertainty
while holding $(u,v)$ fixed. This is the physically primary covariance in the
present formulation because it captures the uncertainty anisotropy generated directly 
by phase measurement and local reconstruction sensitivity.

The second is the \emph{effective full-rank covariance} used in registration.
It is obtained by propagating uncertainty in the local measurement vector
$m=[u,\ v,\ \Phi]^\top$ through the calibrated mapping $h$. This completion preserves
the dominant phase-driven anisotropy while introducing finite transverse uncertainty
for downstream probabilistic methods that require a nonsingular per-point covariance.

\subsection{Depth uncertainty from measured phase precision}
\label{subsec:depth_from_phase}

We first relate phase precision to depth uncertainty using the calibrated
phase-to-depth mapping $z=f(\Phi;u,v)$ introduced in
Section~\ref{subsec:phase_depth_mapping}. At fixed $(u, v)$, first-order propagation
gives
\begin{equation}
\label{eq:cov_sigma_z}
\begin{aligned}
\delta z &\approx
\frac{\partial f}{\partial \Phi}\,\delta\Phi,\\
\sigma_z^2(u,v) &\approx
\left(\frac{\partial f}{\partial \Phi}\right)^2
\sigma_\Phi^2(u,v).
\end{aligned}
\end{equation}

For the affine model in Eq.~\eqref{eq:affine_phase_depth}, $\partial f/\partial\Phi=A$,
and therefore $\sigma_z=|A|\sigma_\Phi$. For the weakly projective model in 
Eq.~\eqref{eq:projective_phase_depth}, the local sensitivity is 

\begin{equation}
\label{eq:cov_projective_df}
\frac{\partial f}{\partial \Phi}=
\frac{A\left(1+\rho(C\Phi+D)\right)-(A\Phi+B)\rho C}
{\left(1+\rho(C\Phi+D)\right)^2},
\end{equation}
where the spatially varying terms are evaluated at the corresponding image position.

Thus, the propagated depth uncertainty is determined by two physically
meaningful quantities: the measured phase variance $\sigma_\Phi^2(u,v)$ and the
local phase-to-depth reconstruction sensitivity $\partial f/\partial \Phi$. Spatial variation in $\sigma_z^2(u,v)$ therefore reflects variation in phase precision, calibrated reconstruction sensitivity, or both, rather than local surface-neighborhood regularization 
\cite{Molimard2013UncertaintyFringeProjection,ODowd2021HeightUncertaintyPDF}.

\subsection{Phase-induced 3D covariance under fixed-pixel reconstruction}
\label{subsec:phase_induced_covariance}

We next propagate phase uncertainty to 3D while treating the image coordinates
$(u,v)$ as fixed. The local sensitivity of the reconstructed point to phase is
\begin{equation}
\label{eq:cov_Jphi_general}
J_\Phi(u,v,\Phi)
=
\frac{\partial h}{\partial \Phi}
=
\frac{\partial X}{\partial z}
\frac{\partial z}{\partial \Phi},
\end{equation}
where the final equality applies when the reconstruction is viewed as the composition
of the phase-depth mapping and a depth-to-3D reconstruction model.

Using the first-order propagation, the phase-induced 3D covariance is
\begin{equation}
\label{eq:cov_sigma_phase_induced}
\Sigma_X^{(\Phi)}(u,v,\Phi)
\approx
J_\Phi(u,v,\Phi)\,
\sigma_\Phi^2(u,v)\,
J_\Phi(u,v,\Phi)^\top.
\end{equation}

This covariance captures the anisotropy induced directly by the phase measurement process. Because a scalar phase perturbation produces a one-dimensional perturbation in
3D under this fixed-pixel model, $\Sigma_X^{(\Phi)}$ is rank-1. In a pinhole-like reconstruction geometry, the dominant direction is approximately aligned with the local viewing ray, producing the characteristic axial-lateral imbalance of structured-light
measurement.

\subsection{Effective full-rank covariance from measurement-space propagation}
\label{subsec:effective_fullrank_covariance}

For downstream probabilistic methods that require a nonsingular per-point
covariance, we augment the phase-induced model by propagating uncertainty in the
local measurement state $m=[u,\ v,\ \Phi]^\top$. Its covariance is modeled as
\begin{equation}
\label{eq:cov_sigma_m}
\Sigma_m(u,v)=
\begin{bmatrix}
\sigma_u^2 & 0 & 0\\
0 & \sigma_v^2 & 0\\
0 & 0 & \sigma_\Phi^2(u,v)
\end{bmatrix},
\end{equation}
where $\sigma_\Phi^2(u,v)$ is measured from repeated phase observations and 
$\sigma_u$, $\sigma_v$ are global effective image-space perturbation scales for transverse uncertainty contributions that are not represented by phase perturbation alone, including finite sampling, interpolation, and residual calibration effects. 
These scales are distinct from the numerical step sizes used to evaluate the Jacobian.
They are estimated once from pooled repeated-plane 3D measurements by fitting the
unexplained transverse repeatability after accounting for the phase-driven rank-1 
contribution. The resulting fitted values are nearly equal, indicating that the
effective lateral perturbation scale is approximately isotropic in the image
plane under the selected operating condition.

The Jacobian of the calibrated mapping with respect to the measurement vector is
\begin{equation}
\label{eq:cov_Jm}
J_m(u,v,\Phi)=
\frac{\partial h}{\partial (u,v,\Phi)}
=
\begin{bmatrix}
\frac{\partial h}{\partial u} &
\frac{\partial h}{\partial v} &
\frac{\partial h}{\partial \Phi}
\end{bmatrix}
\in \mathbb{R}^{3\times3}.
\end{equation}
When an analytic form is not available, the Jacobian is evaluated numerically by
central differences,
\begin{equation}
\label{eq:cov_central_diff}
\begin{aligned}
\frac{\partial h}{\partial u}
&\approx
\frac{h(u+\Delta_u,v,\Phi)-h(u-\Delta_u,v,\Phi)}
     {2\Delta_u},\\
\frac{\partial h}{\partial v}
&\approx
\frac{h(u,v+\Delta_v,\Phi)-h(u,v-\Delta_v,\Phi)}
     {2\Delta_v},\\
\frac{\partial h}{\partial \Phi}
&\approx
\frac{h(u,v,\Phi+\Delta_\Phi)-h(u,v,\Phi-\Delta_\Phi)}
     {2\Delta_\Phi}.
\end{aligned}
\end{equation}

The effective full-rank covariance is written as
\begin{equation}
\label{eq:cov_sigma_effective}
\Sigma_X(u,v,\Phi)
\approx
J_m(u,v,\Phi)\,
\Sigma_m(u,v)\,
J_m(u,v,\Phi)^\top.
\end{equation}

Retaining only the phase term in $\Sigma_m$ recovers the phase-induced covariance
in Eq.~\eqref{eq:cov_sigma_phase_induced}. With nonzero $\sigma_u$ and $\sigma_v$, 
the additional image-space terms provide the transverse completion required for a
practical full-rank covariance, while the dominant scale and direction are still governed by the phase-propagation branch through $\partial h/\partial \Phi$ and
$\sigma_\Phi^2(u,v)$. 

\subsection{SPD regularization and practical construction}
\label{subsec:spd_regularization}

The covariance matrix in Eq.~\eqref{eq:cov_sigma_effective} is symmetric by
construction, but small eigenvalues may arise in practice when the
phase-induced anisotropy is dominant or when local numerical sensitivities are
nearly degenerate. For stable inversion, we apply the eigenvalue floor
\begin{equation}
\label{eq:cov_eigen_floor}
\begin{aligned}
\Sigma_X &= Q\Lambda Q^\top,\\
\Sigma_X^{\mathrm{reg}} &= Q\widetilde{\Lambda}Q^\top,
\qquad
\widetilde{\lambda}_j=\max(\lambda_j,\epsilon).
\end{aligned}
\end{equation}

This regularization is introduced for numerical stability rather than physical modeling, and does not alter the intended physical interpretation of the covariance
for a sufficiently small $\epsilon$ compared to the measured eigenvalues. The resulting regularized field is used in the subsequent experimental analysis.

\section{Experimental Protocol and Implementation}
\label{sec:metrology}

This section summarizes the procedures used to measure phase precision, fit and validate the phase-to-depth mapping, construct the proposed covariance field, and evaluate it in uncertainty-aware registration. Implementation details that do not require separate figures or tables are reported inline.

\subsection{Coordinate conventions, units, and region of interest}
\label{subsec:coords_units}

All reconstructed 3D points and covariances are expressed in the reference
camera coordinate frame $C$ and reported in millimeters. The image
coordinates $(u,v)$ are defined on the undistorted reference-camera grid at
resolution $640\times480$. Unless otherwise stated, depth refers to the
$z$-coordinate in frame $C$.

All phase-variance, mapping-residual, and covariance analyses are restricted to
a valid ROI defined by thresholding fringe modulation and excluding saturated
or otherwise invalid pixels. In the present experiments, the modulation
threshold was fixed at 0.05. The same ROI definition is used throughout unless
otherwise noted.

\subsection{Hardware and imaging configuration}
\label{subsec:hardware}

The structured-light system consists of a monochrome camera (FLIR GS3-U3-32S4M, $2448\times2048$ pixels), and a digital light processing (DLP) projector (Texas Instruments LightCrafter4500, $912\times1140$ pixels). To reduce radial distortion effects, the ROI was limited to the central $640\times480$ pixels as noted. The reference camera defines the coordinate frame $C$ and indexes the absolute phase map $\Phi(u,v)$.

The nominal working distance was $d = 300$ mm, and the calibrated depth range
used in this study was 240--340 mm. Unless otherwise stated, all
phase-variance and phase-to-depth experiments were conducted on a planar
homogeneous target or calibration board placed at multiple depths spanning this
range. The repeated-plane dataset comprised 10 depth positions, 7 tilt
configurations, and 20 repetitions per condition.

To realize the approximately constant-SNR operating condition used throughout
the paper, exposure, projected brightness, optical aperture, and target
material were held fixed across repeated captures. All repeated-plane data were
acquired in a single session under controlled illumination. These settings were
chosen so that fringe modulation remained approximately stable within each
repetition set and over the usable depth range.

\subsection{Camera calibration and undistortion quality}
\label{subsec:camera_calibration}

The reference camera and projector were calibrated using a pinhole model with
planar-target calibration based on an asymmetric circle grid with 147 feature
points. A total of 20 calibration images were acquired over the usable field of
view and working depth range. Calibration quality was evaluated by reprojection
error on the detected target points.

The resulting reference-camera reprojection residuals were 0.043 px mean RMSE,
0.038 px median, and 0.016 px interquartile range, indicating that the camera
calibration was sufficiently accurate for the subsequent undistortion, phase
processing, and reconstruction steps. All subsequent phase processing and
reconstruction were performed on the undistorted reference-camera grid.

\subsection{Acquisition of phase-to-depth calibration data}
\label{subsec:phase_depth_supervision}

To estimate the calibrated phase-to-depth mapping $z=f(\Phi;u,v)$, depth labels
are required on the undistorted reference-camera grid. In this study, these labels are obtained from planar measurements acquired at multiple depths across the calibrated reconstruction range. For each depth position, a planar target was measured and the corresponding absolute phase map $\Phi(u,v)$ was reconstructed on the reference-camera grid. The resulting depth label field is denoted $z_{\mathrm{ref}}(u,v)$ and paired with the measured phase to form calibration samples $(\Phi(u,v), z_{\mathrm{ref}}(u,v))$. Validation was performed using five held-out plane captures not included in fitting.

\subsection{Estimation and validation of the phase-to-depth mapping}
\label{subsec:fitting_protocol}

Using the calibration pairs $(\Phi(u,v), z_{\mathrm{ref}}(u,v))$, we estimate
the fitted phase-to-depth mapping described in
Section~\ref{subsec:phase_depth_mapping}. The fitting model is weakly
projective, with globally shared phase coefficients and smooth spatial terms in
the image coordinates. Its parameters are estimated by nonlinear optimization initialized from a least-squares solution and regularized to enforce smooth, low-amplitude residual structure. During optimization, a hard bound of
$2\,\mu\mathrm{m}$ is imposed on the correction term, although no pixels reach
this bound at convergence. Additional details of the loss design and parameterization are given in\cite{Tak2026SpatiallyCoupled}.

To assess the adequacy of the fitted mapping for uncertainty propagation, we
report residual depth error on both the fitting data and held-out planar
validation data. On the held-out plane(s), the per-pixel residual is computed as 
$e_z(u,v) = z_{\mathrm{est}}(u,v) - z_{\mathrm{ref}}(u,v),$ and summarized by RMSE, 
median, and interquartile range over the valid ROI.

\begin{table}[t]
\centering
\caption{Residual error $e_z$ of the fitted phase-to-depth mapping using a maximum bound of $2\mathrm{\mu m}$. All units are $\mathrm{\mu m}$.}
\label{tab:phase_depth_residuals}
\begin{tabular}{lcccc}
\hline
Dataset & RMSE & Median (IQR) & Max $|\nabla e_z|$ & Max $|e_z|$. \\
\hline
Training & 0.1059 & 0.0668 (0.1527) & 0.2343 & 0.9035 \\
Held-out & 0.1084 & 0.0705 (0.1560) & 0.2023 & 0.8927 \\
\hline
\end{tabular}
\end{table}

\subsection{Phase-precision measurement protocol}
\label{subsec:phase_precision_protocol}

Phase precision is measured from repeated acquisitions of a planar target under
the fixed imaging settings described above. At each depth position, 20 PSP
sequences are captured, and an absolute phase map $\Phi_r(u,v)$ is reconstructed
for each repetition $r$. Over valid pixels, the phase variance is estimated as
\begin{equation}
\sigma_\Phi^2(u,v) = \mathrm{Var}_r\big(\Phi_r(u,v)\big).
\end{equation}

The same repeated captures are also used to estimate the spatial fringe
modulation $I_m(u,v)$ and to verify that the constant-SNR operating condition is
approximately satisfied over the ROI and depth range. The phase-precision
analysis is repeated for $10$ plane positions spanning the calibrated depth
range. 

In addition to the repeated phase measurements used to estimate
$\sigma_\Phi^2(u,v)$, the repeated-plane acquisitions are also used to
estimate the global effective image-space perturbation scales $\sigma_u$ and
$\sigma_v$ for the full-rank covariance model. This estimation is performed
once over all repeated-plane datasets by fitting the residual transverse 3D
repeatability, after subtracting the phase-driven contribution predicted by
the rank-1 covariance branch. The resulting values are then held fixed for
all covariance-field computations and downstream registration experiments.

\subsection{Covariance construction settings}
\label{subsec:covariance_settings}

The proposed sensor-informed covariance was computed as described in
Section~\ref{sec:covariance}. Using the pooled repeated-plane datasets, the
global effective image-space perturbation scales were estimated as
$\sigma_u = 0.0193$ px and $\sigma_v = 0.0193$ px. The two values are nearly
identical, indicating no measurable directional imbalance in the effective
lateral perturbation scale under the selected operating condition. These values
are also close to the independently derived target-floor estimate, and are of the same order as conventional subpixel error scales encountered in calibration and interpolation, although they do not represent the same quantity as camera-calibration reprojection RMSE.

Numerical Jacobians were evaluated by central differences using
$\Delta_u = 0.1$ px, $\Delta_v = 0.1$ px, and
$\Delta_\Phi = 10^{-3}$ in phase units. To ensure stable inversion in
Mahalanobis-weighted registration, a small SPD floor of
$\epsilon = 10^{-10}$ was applied to the covariance eigenvalues. Sensitivity to
these settings is assessed in the covariance characterization results.

\subsection{Registration dataset, G-ICP configuration, and evaluation protocol}
\label{subsec:registration_protocol}

To evaluate the proposed sensor-informed covariance in probabilistic
registration, we acquired a sequence of structured-light scans of a rigid
statue with curved surfaces, local geometric detail, and moderate
self-occlusion under the same imaging settings used in the repeated-plane
experiments. Pairwise registrations were evaluated between adjacent views, with
viewpoint spacing chosen so that neighboring scans retained approximately 85\%
overlap. The acquisition protocol used fixed camera locations established by
physical markers, yielding 14 source--target scan pairs.

Initialization was provided by coarse alignment from the nominal acquisition
geometry. The resulting registration experiment is intended as a downstream
comparison of covariance usefulness under a fixed registration protocol rather
than as a direct validation of covariance accuracy in isolation.

Rigid registration was performed using the G-ICP implementation in the Point
Cloud Library (PCL) in C++. All covariance models used the same point-cloud
preprocessing, correspondence search, and termination criteria. Unless
otherwise stated, the point-clouds were voxel-downsampled with a voxel size of
0.1 mm to approximately $2\times10^4$ points per scan, correspondences were
rejected above a distance threshold of 0.1 mm, and optimization terminated
after 200 iterations or when the update fell below $10^{-4}$ mm. For the
geometry-derived baseline, local neighborhood PCA was computed with $k=20$
neighbors.

Three covariance models were compared:
\begin{enumerate}
    \item Proposed sensor-informed covariance from Section~\ref{sec:covariance},
    \item Constant isotropic covariance $\sigma^2 I$ with $\sigma=0.01$,
    \item Geometry-derived covariance estimated from local neighborhood PCA with
    $k=20$ neighbors.
\end{enumerate}

We report translation RMSE, rotation RMSE, and the median and IQR of both errors. All covariance models use the same initialization and registration parameters; only the covariance assigned to each point is changed. For compact comparison within a single 
pose-error table, both translation and rotation errors are reported in millimeters. Rotational error is converted to a translation-equivalent displacement using the nominal working-distance scale of 300 mm. This conversion is used only for reporting convenience and does not imply that the translation and rotation components are combined into a single 6-DoF pose metric.

\section{Results and Evaluation}
\label{sec:results}

This section evaluates the proposed covariance field in four steps: phase
precision under the selected operating condition, adequacy of the fitted
phase-to-depth mapping for uncertainty propagation, structure of the resulting
covariance field, and downstream performance in rigid registration.

\subsection{Approximately Constant-SNR Verification and Phase Variance}
\label{subsec:results_phase_variance}

\begin{figure*}[t]
\centering
\includegraphics[width=\linewidth]{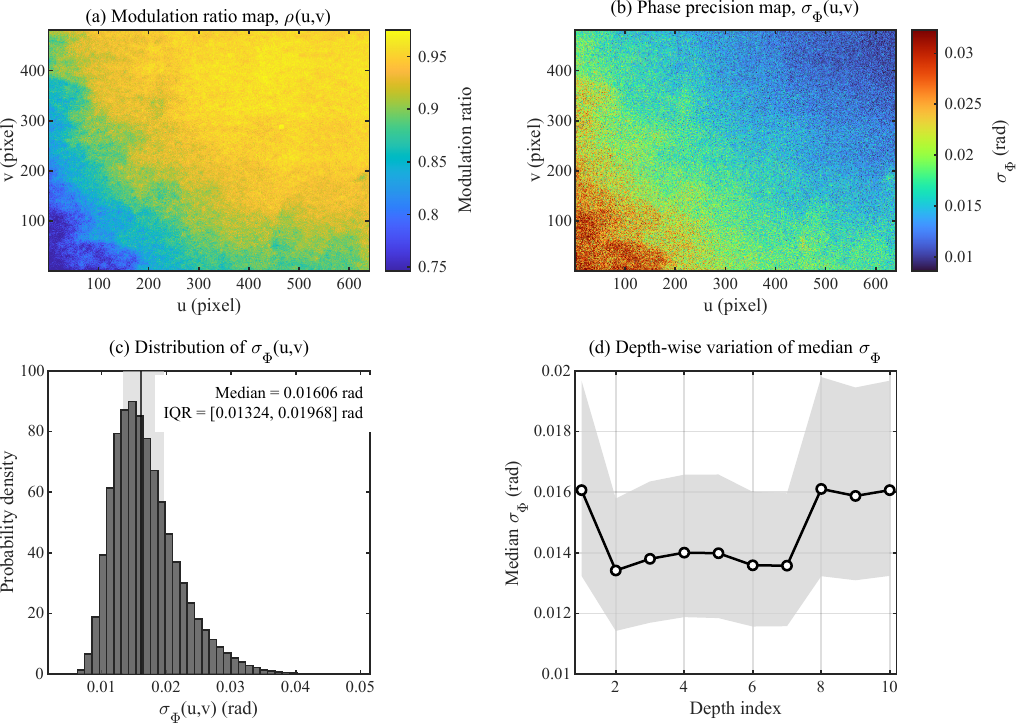}
\caption{Verification of the approximately constant-SNR operating condition and
measured phase precision over the valid ROI and calibrated depth range.
(a) Representative fringe-modulation map $I_m(u,v)$ at a reference depth.
(b) Corresponding map of the measured phase standard deviation
$\sigma_\Phi(u,v)$. (c) Histogram of $\sigma_\Phi(u,v)$ over the valid ROI,
showing the within-ROI distribution of phase precision. (d) Depth-wise median
and interquartile range of $\sigma_\Phi(u,v)$ over the sampled depth positions.
The figure shows that, although modulation is not perfectly uniform over the
image, the measured phase precision remains sufficiently stable across the ROI
and over depth to support the approximately constant-SNR operating condition
used in the covariance model.}
\label{fig:modulation_phasevariance}
\end{figure*}

We first verify that the imaging condition used for phase-precision
measurement is consistent with the intended approximately constant-SNR
operating condition. Figure~\ref{fig:modulation_phasevariance}(a) shows a
representative fringe-modulation map $I_m(u,v)$ within the valid ROI, while
Fig.~\ref{fig:modulation_phasevariance}(b) shows the corresponding map of the
measured phase standard deviation $\sigma_\Phi(u,v)$. The within-ROI
distribution of $\sigma_\Phi(u,v)$ is summarized by the histogram in
Fig.~\ref{fig:modulation_phasevariance}(c), and the depth-wise behavior of the
median and interquartile range is shown in
Fig.~\ref{fig:modulation_phasevariance}(d).

Across the ten sampled depth positions, the median phase standard deviation
remained between 0.0138 and 0.0161 rad, with interquartile ranges between
0.0132 and 0.0197 rad. The valid-ROI fraction remained above 99.99\%
throughout the calibrated depth range. Although the fringe modulation is not
perfectly uniform over the image because of projector geometry, surface
reflectance, and ambient illumination, its spatial pattern remains stable under
fixed acquisition settings. More importantly, the measured phase precision
remains low and varies only modestly across the valid ROI and over the tested
depth range. Together, these results support the use of an approximately
constant-SNR operating condition for the present first-order uncertainty
analysis.

The spatial pattern of the measured phase standard deviation also broadly
follows the modulation pattern in Fig.~\ref{fig:modulation_phasevariance}(a)
and (b), which is consistent with the expected influence of fringe modulation
on phase precision under fixed acquisition settings. This observation is used
here only as an empirical consistency check for the controlled operating
condition; the subsequent covariance model relies on the measured phase
variance itself rather than on modulation as a surrogate quantity.

\subsection{Phase-to-depth mapping residuals and held-out validation}
\label{subsec:results_mapping}

We next evaluate whether the fitted phase-to-depth mapping is sufficiently
accurate to support uncertainty propagation. The corresponding quantitative
summary is reported in Table~\ref{tab:phase_depth_residuals}. Held-out
validation is essential here, because low residual error on planes excluded
from fitting supports the use of the calibrated mapping as the transfer
function in the covariance model over the valid ROI.

In addition to residual magnitude, Table~\ref{tab:phase_depth_residuals} also
reports the maximum residual-gradient magnitude. This is included to check
whether omitted residual-gradient terms could materially affect the local
sensitivity used in uncertainty propagation. In the present formulation,
uncertainty propagation is performed through the fitted smooth phase-to-depth
mapping, whereas residual fitting error is treated as a modeling error term
rather than as part of the nominal transfer function. The reported residual
magnitude and spatial smoothness indicate that this modeling error remains
small relative to the propagated measurement scale, so it is neglected in the
first-order Jacobian model.

\subsection{Covariance field characterization}
\label{subsec:results_covariance}

\begin{figure*}[t]
\centering
\includegraphics[width=\linewidth]{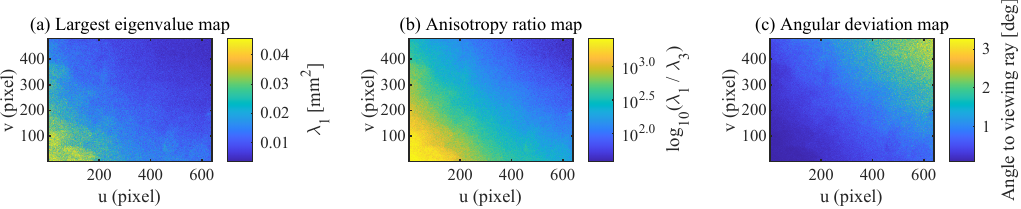}
\caption{Spatial characterization of the proposed covariance field over the
valid ROI. (a) Largest eigenvalue map $\lambda_1(u,v)$, showing the dominant
uncertainty scale. (b) Anisotropy ratio map $\lambda_1(u,v)/\lambda_3(u,v)$,
showing the degree of covariance elongation. (c) Angular deviation between the
dominant eigenvector and the local viewing ray, showing how closely the
principal covariance direction follows the expected reconstruction direction.}
\label{fig:covariance}
\end{figure*}

\begin{figure*}[t]
\centering
\includegraphics[width=\linewidth]{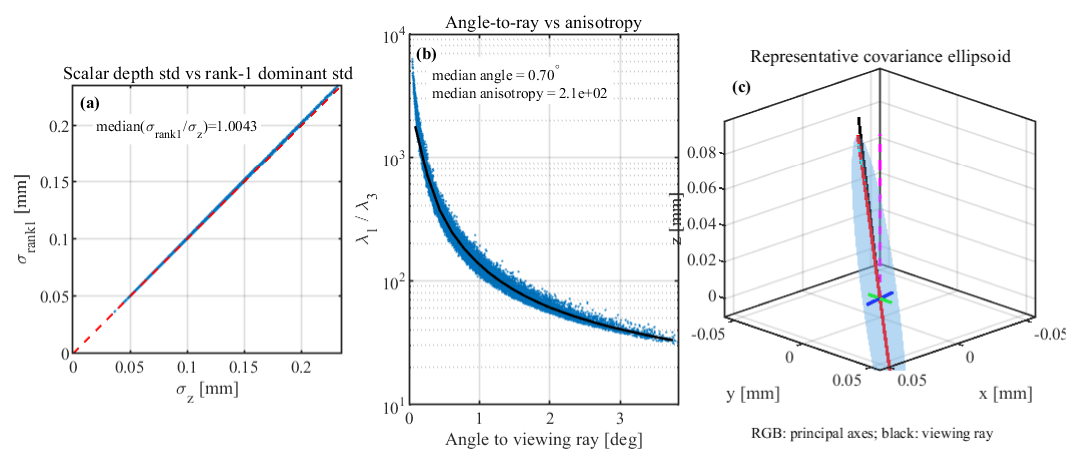}
\caption{Interpretation and consistency checks for the proposed covariance
field. (a) Scalar depth standard deviation versus the dominant standard
deviation of the phase-induced rank-1 covariance, showing agreement of the
phase-driven depth scale. (b) Angle-to-ray versus anisotropy, showing a clear
structural relationship across the valid ROI. (c) Representative covariance
ellipsoid with principal axes, local viewing ray, and scalar uncertainty
comparisons, illustrating how the full-rank covariance extends the phase-driven
model while preserving its dominant ray-like structure.}
\label{fig:covariance_companion}
\end{figure*}

We next evaluate whether the proposed covariance field exhibits the expected
sensor-informed structure. Figure~\ref{fig:covariance} shows the spatial
distribution of three representative covariance summaries: the largest
eigenvalue, the anisotropy ratio, and the angular deviation between the
dominant eigenvector and the local viewing direction. Figure
\ref{fig:covariance_companion} provides two complementary consistency checks:
agreement between scalar depth standard deviation and the dominant standard
deviation of the phase-induced rank-1 covariance, and the relationship between
anisotropy and angle to the viewing ray. It also shows one representative
covariance ellipsoid together with the local viewing ray and scalar uncertainty
comparisons.

The quantitative summary of the covariance field is reported in
Table~\ref{tab:covariance_summary}. The dominant eigenvalue remains
substantially larger than the smallest eigenvalue throughout the valid ROI,
confirming that the proposed covariance field is strongly anisotropic. At the
same time, the angle between the dominant eigenvector and the local viewing ray
remains small, indicating that the principal covariance direction stays close
to the expected reconstruction direction.

Taken together, the spatial maps, the summary statistics, and the companion
consistency checks support three key observations. First, the dominant covariance
scale remains consistent with the scalar depth uncertainty predicted from phase
propagation, indicating that the rank-1 branch preserves the physically
meaningful range scale. Second, the covariance field remains strongly
anisotropic over the valid ROI. Third, the dominant axis remains closely
aligned with the local viewing direction, showing that the effective full-rank
completion preserves the sensor-informed ray-like structure. These results therefore support the internal consistency of the proposed covariance construction.

\begin{table}[t]
\centering
\caption{Summary statistics of the covariance field over the valid ROI.
Eigenvalues are reported in mm$^2$, and angular deviation is reported in
degrees.}
\label{tab:covariance_summary}
\begin{tabular}{lccc}
\hline
Quantity & Mean & Median & IQR \\
\hline
Largest eigenvalue $\lambda_1$ & 0.0137 & 0.0113 & 0.0091 \\
Smallest eigenvalue $\lambda_3$ & 5.72e-5 & 5.51e-5 & 3.55e-5 \\
Anisotropy ratio $\lambda_1/\lambda_3$ & 383.0 & 205.5& 311.3 \\
Angle to local viewing ray & 0.90 & 0.70 & 0.80 \\
\hline
\end{tabular}
\end{table}

\subsection{Registration accuracy comparison}
\label{subsec:results_registration}

Finally, Table~\ref{tab:registration_results} summarizes G-ICP registration
accuracy over the 14 source--target pairs. The constant isotropic covariance
produces the largest translation and rotation errors, while the conventional
geometry-derived covariance gives the lowest errors on the present dataset. The
proposed sensor-informed covariance lies between these baselines and improves
over the isotropic model in both RMSE and median error. The implications of this
difference are discussed in Section~\ref{sec:discussion}.

\begin{table}[t]
\centering
\caption{Rigid registration results for G-ICP under different covariance models.}
\label{tab:registration_results}
\footnotesize
\begin{tabular}{lcccc}
\hline
Model & T-RMSE & T-Med. (IQR) & R-RMSE & R-Med. (IQR)  \\
\hline
Isotropic & 0.7176 & 0.2061 (0.2132) & 1.1340 & 1.257 (0.3021) \\
Geometric & 0.1097 & 0.0887 (0.1023) & 0.3516 & 0.3023 (0.1061) \\
Proposed & 0.2135 & 0.1441 (0.1535) & 0.4216 & 0.4051 (0.1857) \\
\hline
\end{tabular}
\end{table}

\section{Discussion}
\label{sec:discussion}

The experiments support the proposed covariance model at two complementary levels. First, the phase-driven component is grounded directly in repeated phase measurements 
and the calibrated smooth phase-to-depth mapping. Phase precision remains stable under 
the selected operating condition, the fitted mapping exhibits low held-out residual
error, and the dominant standard deviation of the rank-1 phase-induced covariance
agrees with the scalar depth uncertainty predicted from phase propagation. 
Together with the observed ray alignment, these results support the physical interpretation of the phase-induced component as the dominant uncertainty direction of the present structured-light system.

Second, the effective full-rank covariance extends this phase-induced model with global
image-space perturbation scales $\sigma_u$ and $\sigma_v$. They should be interpreted as
effective lateral scales rather than independently identified physical noise sources.
The diagonal form of $\Sigma_m$ likewise constitutes a first-order approximation in which cross-correlations among $u$, $v$, and $\Phi$ are neglected.

The present experiments establish the internal consistency and practical plausibility
of the resulting covariance field rather than a complete multivariate validation against
independent 3D perturbation ground truth. Calibration-parameter uncertainty,
undistortion uncertainty, phase-to-depth model uncertainty, and inter-pixel error 
correlation are not propagated explicitly, and the model is evaluated under the 
controlled imaging regime used to estimate its phase and lateral precision. Extending
this model to spatially varying operating conditions and explicitly propagating these
additional uncertainty sources would provide a more complete error description.

The G-ICP experiment provides a downstream comparison of covariance utility. The 
proposed covariance improves substantially over a constant isotropic model but remains
less accurate than the conventional geometry-derived covariance on the present
dataset. This difference is consistent with the distinct information represented by
the two constructions: the proposed covariance describes sensor-informed measurement
uncertainty, whereas the geometry-derived covariance directly encodes local surface
constraints that are particularly favorable for G-ICP. Combining these complementary
sources of information in a hybrid sensor-geometry model is a natural direction for
future work.

\section{Conclusion}
This paper presented a sensor-informed first-order method for constructing a
per-point $3\times3$ covariance field for structured-light 3D reconstruction
from measured phase precision and calibrated phase-to-depth and phase-to-3D
mappings. In contrast to conventional geometry-derived covariance models, the
proposed formulation does not depend on local surface neighborhoods and remains
directly interpretable in terms of phase precision, reconstruction sensitivity,
and imaging geometry. The formulation distinguishes a physically primary
phase-induced covariance from a practical full-rank completion used for
probabilistic registration.

Under an approximately constant-SNR operating condition, the experiments showed
that phase precision can be measured reproducibly over the working range and
that the fitted phase-to-depth mapping is accurate enough for first-order
propagation. The resulting covariance field exhibited the expected anisotropic
structure of structured-light measurement, including strong axial--lateral
imbalance, close alignment of the dominant axis with the local viewing
direction, and consistency between the rank-1 phase-induced scale and scalar
depth uncertainty.

In G-ICP registration, the proposed covariance provided a clear improvement
over a constant isotropic model, but remained less accurate than a
geometry-derived covariance on the present dataset. This outcome is consistent
with the different roles of the two models: the proposed covariance represents
sensor-informed measurement uncertainty, whereas the geometry-derived covariance
encodes local surface constraint that is especially favorable for G-ICP. Future
work will extend the formulation to spatially varying SNR, explicit treatment
of spatial correlation and model uncertainty, stronger covariance-validation
experiments, and hybrid sensor--geometry covariance models.

\section*{Acknowledgements}
This research was supported by the Culture, Sports and Tourism R\&D Program through the Korea Creative Content Agency grant funded by the Ministry of Culture, Sports and Tourism in 2024 (Project Name: Global Talent for Generative AI Copyright Infringement and Copyright Theft, Project Number: RS-2024-00398413, Contribution Rate: 95\%), Yonsei University Research Fund (Project Number: 2024-22-0483, Contribution Rate: 4\%), and International Joint Research Grant by Yonsei Graduate School (Contribution Rate: 1\%).

\bibliography{ref}
\bibliographystyle{IEEEtran}

\vfill

\end{document}